\documentclass[times,onecolumn,authoryear]{elsarticle}

\usepackage{prletters}
\usepackage{framed,multirow}

\usepackage{amssymb}
\usepackage{latexsym}

\usepackage{url}
\usepackage{xcolor}
\usepackage{multirow}
\usepackage{mathtools}
\usepackage{makecell}
\usepackage{graphicx}
\usepackage{nicefrac}

\definecolor{newcolor}{rgb}{.8,.349,.1}

\journal{Pattern Recognition Letters}

\begin{document}

\title{SM-DTW: Stability Modulated Dynamic Time Warping for signature verification}

\author[1]{Antonio Parziale\corref{cor1}} 
\cortext[cor1]{Corresponding author: 
  Tel.: +39-089-96-4177;}
\ead{anparziale@unisa.it}
\author[2]{Moises Diaz}
\ead{mdiazc@unidam.es}
\author[3]{Miguel A. Ferrer}
\ead{miguelangel.ferrer@ulpgc.es}
\author[1]{Angelo Marcelli}
\ead{amarcelli@unisa.it}

\address[1]{DIEM, University of Salerno, 84084 Fisciano (SA), Italy}
\address[2]{Universidad del Atlantico Medio, Las Palmas de Gran Canaria, 35017, Spain}
\address[3]{iDeTIC. Universidad de Las Palmas de Gran Canaria, 35017, Spain}

\begin{abstract}
Building upon findings in computational model of handwriting learning and execution, we introduce the concept of stability to explain the difference between the actual movements performed during multiple execution of the subject's signature, and conjecture that the most stable parts of the signature should play a paramount role in evaluating the similarity between a questioned signature and the reference ones during signature verification. We then introduce the Stability Modulated Dynamic Time Warping algorithm for incorporating the stability regions, i.e. the most similar parts between two signatures, into the distance measure between a pair of signatures computed by the Dynamic Time Warping for signature verification. Experiments were conducted on two datasets largely adopted for performance evaluation. Experimental results show that the proposed algorithm improves the performance of the baseline system and compares favourably with other top performing signature verification systems.
\end{abstract}


\noindent{\Large SM-DTW: Stability Modulated Dynamic Time Warping for signature verification}

\noindent{Antonio Parziale, Moises Diaz, Miguel A. Ferrer, Angelo Marcelli}

\section*{Abstract}
\noindent Building upon findings in computational model of handwriting learning and execution, we introduce the concept of stability to explain the difference between the actual movements performed during multiple execution of the subject's signature, and conjecture that the most stable parts of the signature should play a paramount role in evaluating the similarity between a questioned signature and the reference ones during signature verification. We then introduce the Stability Modulated Dynamic Time Warping algorithm for incorporating the stability regions, i.e. the most similar parts between two signatures, into the distance measure between a pair of signatures computed by the Dynamic Time Warping for signature verification. Experiments were conducted on two datasets largely adopted for performance evaluation. Experimental results show that the proposed algorithm improves the performance of the baseline system and compares favourably with other top performing signature verification systems.


\section{Introduction}
Dynamic Time Warping (DTW) is an algorithm largely used in automatic signature verification (ASV) to detect forgeries. Indeed, some of the most performing ASV systems in the current literature are based on such algorithm. DTW's elastic property is likely the key of its success. Such property allows to enlarge and shorten a cost function in order to find the best elastic matching between signals. 

Some DTW-based systems apply some normalizations to the obtained dissimilarity measure, e.g. \cite{diaz2016dynamic,morales2017signature,fischer2015robust,kholmatov2009susig}. Other authors work out the optimization matrix because further information can be extracted~\cite{sharma2017exploration}. Moreover, in combination with vector quantization (VQ), which models the signature through prototype vectors, the DTW seems to be competitive as well~\cite{sharma2016enhanced}. Similarly, DTW has shown its power when signatures were represented by features derived from a Gaussian Mixture Model (GMM)~\cite{sharma2017novel} or from the kinematic theory of rapid movements~\cite{fischer2017signature, gomez2015enhanced}. In the stability field, the DTW algorithm has been used together with the Direct Matching Points (DMP) in order to identify the most local stable features of a signature~\cite{pirlo2015multidomain} or for the stability and complexity analysis of dynamic signatures~\cite{pirlo2015stability}.

However, not only dynamic signature systems, but also off-line ASVs have used DTW~\cite{shanker2007off}. As an example, the distance of strokes extracted from image-based signature is measured by means of DTW~\cite{ye2005off}.

Signatures are complex movements composed by elementary movements, or strokes, which are concatenated in such a way that their execution produces the desired trajectory with the minimum amount of metabolic energy. Along the years and by repeated practice, such an optimization is learned and the signature becomes a distinctive feature of the subject. The effect of the learning is stored in the brain as a motor plan that incorporates both the sequence of target points, i.e. the points the strokes aim to reach, and the sequence of motor commands to be executed to draw the desired shape between them \cite{Senatore2012neural}. The former encoding is effector independent, while the latter depends on the effector selected for signing \cite{Raibert1977}.

Multiple executions of the signature produced by the same effector, therefore, should produce the same trajectory, as they are the execution of the same motor plan executed by the same effector. The actual configuration of the effector, however, may change from one execution to another, because of the psychophysical conditions of the subject. During the execution of the movement, thus, the spinal cord may modulate the commands of the motor plan in reaction to the visual and proprioceptive feedbacks in receives, representing the current state of the effector, in order to keep the actual trajectory as close as the intended one \cite{Marcelli2013SomeOO,Parziale2015,ParzialeTesi,Ferrer2017}. Because of its nature and intended aims, such a modulation should produce small variations of the shape along the trajectory. 

On the other hand, as the complexity of the signing movements increases, so does the complexity of the motor plan and the amount of metabolic energy required for its execution. Thus, during the learning, the subject may find more convenient to break the entire movements into sub-movements, for each of whom its motor plan is stored. During the execution, the motor plans are retrieved, as well as their temporal order, and executed, each one like an elementary movement. Between two successive ones, however, the movement to perform depends on the relative position between the \textit {actual} position of pentip at the end of the first movement and the \textit {intended} position of the pentip for starting the second movement. As such movements are not stored in the motor plan, they are computed on the fly, by resorting to the repertoire of motor plans the subject has previously learned. During their execution, therefore, it is very likely that a bigger modulation of the commands by the spinal cord will be needed to adjust the movements towards to intended target. Larger modulation, in turn, may produce larger effect on the shape of the trajectory corresponding to those movements. 

According to these observations, we conjecture that the signing habits of a subject have been encoded in the motor plan that has been learned, and thus during multiple executions of the signature, the movements corresponding to the commands stored into the motor plan should produce very similar shapes, while those computed on the fly may exhibit a larger variability.
As a consequence, the trajectory of a signature can be thought as composed of stability regions, originated by the execution of the motor plan, and other regions, originated by the execution computed on the fly. Such stability regions, thus, convey the most relevant information about the signing habits, and therefore should be given more importance during the verification.

The idea of signature stability has been exploited by different algorithms that have been grouped as model-based, feature-based and data-based approaches \cite{pirlo2015stability}. Model-based approaches build a model of the writer, for example by using Hidden Markov Model \cite{salicetti2008client}, and evaluate stability in term of variability of model parameters. Feature-based approaches estimate signature stability by extracting and matching feature vectors \cite{LEI20052483}. Data-based approaches use raw-data for matching signatures and estimating their differences. \cite{pirlo2013verification} and \cite{diaz2015off} use optical flow for evaluating the deformation process that transforms a genuine signature in another, while \cite{impedovo2012analysis} use DTW to compare genuine signatures to identify matching points that can indicate the presence of a stable region of the signature. 

The approach adopted in this paper differs from the others because it represents explicitly the motor plans in terms of stability regions and uses this information to evaluate the similarity between the signatures. In this framework, we introduce the Stability Modulated Dynamic Time Warping (SM-DTW) algorithm, which aims at incorporating into the computation of the DTW the role of the stability regions by giving larger weights to the shape difference between stability regions than in case of ordinary regions. 

The remaining of this paper is organized as it follows. Section 2 introduces the SM-DTW algorithm main components, namely the detection of the stability regions, the estimation of the relevance of the stability regions in capturing the signing habits of a subject and the modulation of the DTW computation based on the relevance of the stability regions. Section 3 describes the experiments performed and reports the results obtained on two signature datasets largely adopted for evaluating the performance of signature verification systems. Eventually, Section 4 discusses to which extent the experimental results support our conjecture and the future investigations they suggest.

\section{The SM-DTW algorithm}
\label{regions}  
The proposed algorithm is a weighted version of the classical DTW algorithm that exploits the detection of stability regions between a set of $N$ reference signatures $\mathbf{R}=\{R_{1},...,R_{N}\}$ generated by a subject and a questioned signature $Q$.

In the light of the theoretical framework described above, stability regions are detected by representing each signature by the motor plan it derives from. As this last is encoded by a sequence of strokes, they have to be retrieved from the learned sequence. For this purpose, any of the stroke segmentation methods proposed in literature can be adopted. In this work we have used the multiresolution algorithm described in \cite{DeStefano2004}, where the desired segmentation points are the points corresponding to the most salient changes in curvature. We have also considered to perform this step by using the algorithm proposed in \cite{OREILLY2009}, which exploits dynamic information, but experiments have shown that it tends to segment multiple executions of the same motor plan into a different number of strokes \cite{parisi2018}, and thus does not allow to reliably infer the stability regions as it is described next.

After segmentation, therefore, both questioned and reference signatures are represented by sequences of strokes, as in eq. \ref{seqstroke} and \ref{seqstroker}:

\begin{equation}
Q=\{\tilde{q}^{1},\dots,\tilde{q}^{j},\dots\tilde{q}^{z}\}
\label{seqstroke}
\end{equation}  

\begin{equation}
R_{i}=\{\tilde{r}^{1}_{i},\dots,\tilde{r}^{j}_{i},\dots,\tilde{r}^{c_{i}}_{i}\}
\label{seqstroker}
\end{equation}
where $\tilde{q}^{j}$ is the $\mathit{j^{th}}$ stroke of $Q$ and $\tilde{r}^{j}_{i}$ is the $\mathit{j^{th}}$ stroke of the $\mathit{i^{th}}$  reference $R_{i}$ belonging to $\mathbf{R}$.

\subsection{Searching for the stability regions}
According to the definition given in the Introduction, the stability regions between two signatures are the longest sequences of strokes having similar shapes and aiming at reaching the same target points. In \cite{Parziale2014} a method for finding the stability regions between two signatures based upon this definition has been presented, and will be summarized in the following. In order to find the longest sequences of strokes having similar shapes, the trajectory is represented by the digital line obtained by the Bresenham algorithm \cite{bresenham1965algorithm} to the temporal sequence of x-y coordinates provided by the tablet. To find stability regions a matching algorithm that exploits the concept of saliency proposed to account for attentional gaze shift in primate visual system is adopted \cite{Itti1998}. 

Given two signatures $R_{i}$ and $Q$, segmented in $c_{i}$ and $z$ strokes respectively, the method builds a multiscale representation of the trajectory made by $H=min(c_{i},z)$ scales. At each scale $h$, where $h \in \{1, 2,\dots, H\}$, the method measures the similarity between all the sequences made by $h$ strokes and computes the similarity matrix $\mathrm{SD}^{h}$, made by $c_{i}\times z$ elements. At the lowest scale ($h=1$), the value of the $(y^{th},l^{th})$ element of the similarity matrix $\mathrm{SD}^{1}$ is equal to the shape similarity between the pair of strokes $(\tilde{r}^{y}_{i},\tilde{q}^{l})$, which is computed by adopting the Weighted Edit Distance (WED) presented in \cite{DeStefano2005v2}. This distance varies between $0$ and $100$ and it is equal to $0$ when two strokes with the same shape but opposite direction are matched and to $100$ when two identical strokes are compared. 
When $h>1$, the $(y^{th},l^{th})$ element of the similarity matrix $\mathrm{SD}^{h}$ is set equal to the mean similarity of the sequence composed of $h$ pair of strokes, as in eq. \ref{eqshapesimilarity}:

\begin{equation}
	\mathrm{SD}^{h}(y,l)=\begin{cases}	
		\dfrac{\sum_{p=0}^{h-1}\mathrm{SD}^{1}(y+p,l+p)}{h},  & \mbox{if}\;\mbox{$(y+h-1<c_{i})$}\;\;\\ 
		& \mbox{and $(l+h-1<z)$} \\
		0, & \mbox{otherwise}\\
	\end{cases}
	\label{eqshapesimilarity}
\end{equation}
where $\mathrm{SD}^{1}(y+p,l+p)=\mathrm{WED}(\tilde{r}^{y+p}_{i},\tilde{q}^{l+p})$, $0\leq y\leq c_{i}-1$ and $0\leq l\leq z-1$.

The most salient sequences of strokes, i.e. the ones with a similarity greater than a threshold $th_{LCS}$, are selected at each scale and are eventually combined in order to build a saliency map. In particular, at each scale, the similarity of a pair of stroke is set equal to 0 if its value is below $th_{LCS}$. 

The saliency map $\mathrm{SAM}$ is an $c_{i}\times z$ matrix whose elements report, for each pair of strokes, the maximum of their similarity across the scales at which the pair of strokes belongs to one of the most salient sequences, as in eq. \ref{equazioneMAP}:

\begin{equation}
\mathrm{SAM}(y,l)=\max(\mathrm{SD}^{1}(y,l),\dots,\mathrm{SD}^{h}(y,l),\dots,\mathrm{SD}^{H}(y,l))
\label{equazioneMAP}
\end{equation}
where $0\leq y\leq c_{i}-1$ and $0\leq l\leq z-1$.

The longest sequences of saliency map elements with a value greater than zero represent the longest sequences of strokes having similar shape between the two signatures and from here on are denoted by \textit{LSSSs}.
None, one or many \textit{LSSSs} can be detected, respectively when two signatures completely different, completely equal or similar only in some parts are compared. In general, $\textit{LSSSs}_{R_{i}Q}=\{\textit{LSSS}^{1}_{R_{i}Q},\dots,\textit{LSSS}^{k}_{R_{i}Q}\}$, and each $\textit{LSSS}$ is a sequence of matching strokes

\begin{equation}
\textit{LSSS}^{j}_{R_{i}Q}=\{S_{R_{i}},S_{Q}\}=\{(\tilde{r}^{y}_{i},\tilde{q}^{l}),\dots,(\tilde{r}^{u}_{i},\tilde{q}^{o})\}
\end{equation} 
where $S_{R_{i}}=\{\tilde{r}^{y}_{i},\dots,\tilde{r}^{u}_{i}\}$ and $S_{Q}=\{\tilde{q}^{l},\dots,\tilde{q}^{o}\}$ are sequences of \textit{ns} matching strokes. 

Once the \textit{LSSSs} have been found, the method evaluates whether the matching sequences correspond to a stability region by estimating if $\mathit{S_{R_{i}}}$ and $\mathit{S_{Q}}$ aim at reaching the same sequence of target points. We must bear in mind that the absolute position of the target points may be changed voluntarily by the subject by changing the size of the signing movements. However, such a change does not affect the relative position among the target points. Thus, for establishing whether the two sequences $\mathit{S_{R_{i}}}$ and $\mathit{S_{Q}}$ aim at reaching the same target points, the starting and the ending point of each stroke are joined by a segment and the angles between successive segments are computed. Therefore, the stroke sequences $\mathit{S_{R_{i}}}$ and $\mathit{S_{Q}}$ are represented by two sequences of \textit{ns-1} angles, $\alpha_{S_{R_{i}}}=\{\alpha_{1},\dots,\alpha_{ns-1}\}$ and $\beta_{S_{Q}}=\{\beta_{1},\dots,\beta_{ns-1}\}$ respectively. Eventually, the similarity vector of the sequence of target points of $\mathit{S_{R_{i}}}$ and $\mathit{S_{Q}}$, denoted by $T(\mathit{S_{R_{i}}},\mathit{S_{Q})}$, is computed by exploiting the idea that the more similar the angles, the closer the relative position of the target points aimed by the strokes of $\mathit{S_{R_{i}}}$ and $\mathit{S_{Q}}$, as in eq. \ref{eqgoodness}:

\begin{equation}\label{eqgoodness}
	T(\mathit{S_{R_{i}}},\mathit{S_{Q}})= [\mathrm{cos}(2\Delta\gamma_{1}),...,\mathrm{cos}(2\Delta\gamma_{ns-1}),1]
\end{equation}
where
\begin{equation}\label{eqdiffangoli}
\Delta\gamma(\mathit{S_{R_{i}}},\mathit{S_{Q}})=\alpha_{\mathit{S_{R_{i}}}}-\beta_{\mathit{S_{Q}}}
\end{equation}

The shape similarity measure and the target point position similarity measure are combined into the \emph{global similarity measure} (\textit{GS}) defined as in eq. \ref{eqglobal}:
\begin{equation}\label{eqglobal}
GS=\sum_{k=1}^{ns}\mathrm{SAM}(\mathit{S_{R_{i}}}(k),\mathit{S_{Q}}(k))*T(\mathit{S_{R_{i}}}(k),\mathit{S_{Q}}(k))
\end{equation}

Eventually, we consider that the motor plan of a signature must be fairly complex: too simple motor plans, routinely used by the subject to encode simple shapes, e.g. the loop on top of the letter e, h, l and similar may be part of the repertoire of motor plans every subject has learned by practicing handwriting, and therefore are not a specific feature of each subject. We assume as measure of the complexity of the motor plan the number of elementary movements it contains. Thus, the desired stability regions are the \textit{LSSSs} longer than the threshold $th_{LEN}$ and whose global similarity is greater than the threshold $th_{GS}$, and from here on are denoted as $\textit{LSSSs}^{*}$.

At the end of this step, therefore, each point sampled by the tablet is labelled depending on whether or not it belongs to stability regions.

\subsection{Estimating the relevance of stability regions}
In the previous subsection we have presented a method for finding stability regions when a signature $Q$ is compared with a single reference signature $R_{i}$. In way of principle, one would expect that the same stability regions appear when the comparison is performed respect to other signatures written by the same subject. In practice, many sources of variability influence the execution of a complex movement and differences can be observed in many replications of a signature. In particular, different executions of a signature encoded by more than one motor program may produce different traces due to the variability in movements executed for connecting the instances of two consecutive motor programs. If more than one reference signature are available, it is possible to better modelling the signing habits of a writer, i.e. to infer his/her motor programs, because we can take into account the effects of writer's motor variability. 

Given a set $\mathbf{R}$ of $N$ reference signatures written by a subject, it is possible to evaluate which parts of the signature $Q$ are more representative of subject's signing habits being less affected by motor variability by detecting the stability regions between $Q$ and each of the references. The set of all stability regions found by comparing $Q$ with all the signatures in $\mathbf{R}$ is denoted as:

\begin{equation}
\mathit{StReg}(\mathbf{R},Q)=\{\textit{LSSSs}^{*}_{R_{1}Q},\dots,\textit{LSSSs}^{*}_{R_{N}Q}\}
\end{equation}

By evaluating the stability of $Q$ respect to each reference signature, we count how many times each stroke of $Q$ belongs to a stability region. From here on, we denote with $\mathit{score(Q)}$ the set of counters associated with each stroke of $Q$:

\begin{equation}
\mathit{relevance}(\tilde{q}^{j})=\sum_{k=1}^{N}g(\textit{LSSSs}^{*}_{R_{k}Q},\tilde{q}^{j})
\end{equation}

\begin{equation}
g(\textit{LSSSs}^{*}_{R_{k}Q},\tilde{q}^{j})=\begin{cases}
1,  & \mbox{if } \mbox{	$\tilde{q}^{j}\in \textit{LSSSs}^{*}_{R_{k}Q} $ } \\
0, & \mbox{otherwise }\\
\end{cases}
\end{equation}

\begin{equation}
	\mathit{score(Q)}=[\mathit{relevance}(\tilde{q}^{1}),\dots,\mathit{relevance}(\tilde{q}^{z})]
\end{equation}

The more times a stroke belong to a stability region, the more it is representative of the writer signing habit, the more it is relevant in the evaluation of genuineness. The counter associated to a stroke varies between $0$ and $N$, depending on whether the stroke does not belong to any stability region or it is included in all the $\textit{LSSSs}^{*}$.

\subsection{Modulating DTW computation by stability regions}
To evaluate the dissimilarity between a reference signature represented by $m$ sampled points, $R_{i}=\{r^{1}_{i},\dots, r^{m}_{i}\}$, and a questioned signature represented by $n$ sampled points, $Q=\{q^{1},\dots,q^{n}\}$, we have used a simple design of Dynamic Time Warping algorithm, which is able to align two non-linear temporal sequences through dynamic programming. 

Each point of a signature was represented by a feature vector $\mathcal{F(\cdot)}$ made by $f$ features, therefore the two signatures were compared in a $\Re^{f}$ space. In particular, we denote with $\overline{R_{i}}$ the $m\times f$ feature matrix representing signature $R_{i}$ and with $\overline{Q}$ the $n\times f$ feature matrix representing signature $Q$.

The classical DTW implementation compares the two feature matrices by creating a $m\times n$ cost matrix whose $(k^{th},j^{th})$ element is equal to the euclidean distance between $\mathcal{F}(r^{k}_{i})$ and $\mathcal{F}(q^{j})$, computed as in eq. \ref{eqdtwdistance}.

\begin{equation}
\label{eqdtwdistance}
d(\mathcal{F}(r^{k}_{i}),\mathcal{F}(q^{j}))=\parallel(\mathcal{F}(r^{k}_{i}),\mathcal{F}(q^{j}))\parallel
\end{equation}

The best alignment between the two matrices is the one with the lowest distance $d\left(\overline{Q}, \overline{R}_{i} \right)$ that can be found by recursively applying the eq. \ref{eqmindtw}: 

\begin{equation}
\label{eqmindtw} 
\psi(k,j)=d(\mathcal{F}(r^{k}_{i}),\mathcal{F}(q^{j}))+\min
\begin{Bmatrix}\psi(k,j-1),\\ \psi(k-1,j-1),\\ \psi(k-1,j) \end{Bmatrix}
\end{equation}
where $\psi(k,j)$ is the cumulative distance up to the $(k^{th},j^{th})$ element. In particular, the distance between the two signatures is computed as in eq. \ref{distancepath}:

\begin{equation}
\label{distancepath}
d\left(\overline{Q}, \overline{R}_{i} \right) = \min_p \displaystyle \sum_{k=1}^{|p|}d(\mathcal{F}(r^{p_{k,r}}_{i}),\mathcal{F}(q^{p_{k,q}})) 
\end{equation}  
where $p=\{(p_{1,r},p_{1,q}),\dots,(p_{|p|,r},p_{|p|,q})\}$ is the minimum warping path between the two feature matrices and $|p|$ is the path length.

The classical implementation weights uniformly the distances between all points of the two series. Instead, in \cite{Jeong2011} the authors introduced the concept of Weighted Dynamic Time Warping (WDTW), a penalty-based DTW that holds in consideration phase differences between reference and testing points. 

The proposed SM-DTW is a specialisation of the WDTW for the signature verification domain and it exploits the algorithm for the detection of stability regions described before. In particular, the distance between a reference and a questioned point is weighted according to the representativeness of the stroke the questioned point belongs to and to the distance between the two feature vectors representing the two points. By applying the SM-DTW algorithm, the $(k^{th},j^{th})$ element of the cost matrix is computed as in eq. \ref{eqwdtwdistance}.

\begin{equation}
\label{eqwdtwdistance}
d^{\text{SM}}(\mathcal{F}(r^{k}_{i}),\mathcal{F}(q^{j}))= w(r^{k}_{i},q^{j})*\parallel(\mathcal{F}(r^{k}_{i}),\mathcal{F}(q^{j}))\parallel
\end{equation}
In order to properly penalize points that are outside the stability regions but have a distance $d$ equal to zero, their distance is set equal to the minimum of the cost matrix.

Eventually, the dissimilarity between the two signatures is computed as in eq. \ref{distancepathsm}.

\begin{equation}
\label{distancepathsm}
d^{\text{SM}}\left(\overline{Q}, \overline{R}_{i} \right) = \min_p \displaystyle \sum_{k=1}^{|p|}d^{\text{SM}}(\mathcal{F}(r^{p_{k,r}}_{i}),\mathcal{F}(q^{p_{k,q}})) 
\end{equation}  
The weight $w(r^{k}_{i},q^{j})$ is computed taking into account if points $r^{k}_{i}$ and $q^{j}$ belong to strokes included in a stability region or not.

\begin{equation}
\label{pesi}
w(r^{i}_{k},q^{j})=
\textit{sigm} (d(\mathcal{F}(r^{k}_{i}),\mathcal{F}(q^{j})),b,c)
\end{equation}

where
\begin{equation}
	\label{sigmoideS}
\textit{sigm}(d(\cdot,\cdot),b,c)=1+\dfrac{1}{1+\exp (-c*(d(\cdot,\cdot)-b)}
\end{equation}

The two parameters $b$ and $c$ depend on the relevance of the stroke the questioned point belong to. In particular, given a stroke $\tilde{q}^{j}$ with relevance $\mathit{relevance}(\tilde{q}^{j})$, the parameters $b$ and $c$ are computed as in eq. \ref{parameterB} and \ref{parameterC}, respectively.

\begin{equation}
	\label{parameterB}
	b=\begin{cases}
		b_{0},  & \;  \mbox{if }\mathit{relevance}(\tilde{q}^{j})=0\;\;\\
		b_{max}+\frac{(b_{min}-b_{max})*(\mathit{relevance}(\tilde{q}^{j})-1)}{N-1}& \;  \mbox{otherwise }\;\;\\
	\end{cases}
\end{equation}

\begin{equation}
\label{parameterC}
c=\begin{cases}
c_{0},  & \;  \mbox{if }\mathit{relevance}(\tilde{q}^{j})=0\;\;\\
c_{min}+\frac{(c_{max}-c_{min})*(\mathit{relevance}(\tilde{q}^{j})-1)}{N-1}& \;  \mbox{otherwise }\;\;\\
\end{cases}
\end{equation}

By using equations (\ref{eqwdtwdistance}-\ref{parameterC}), SM-DTW modulates the distance between two points according to the relevance of the strokes they belong to. It is worth noting that by adopting this kind of modulation the dissimilarity between two points is accentuated as the relevance of the stroke increases: the bigger the relevance (i.e the bigger the stability), the bigger the weight. The rationale behind this choice is that, given the definition of stability, no dissimilarities are expected between part of the reference and part of the questioned belonging to the stability region. So, if they occur, the distance between the corresponding points should be enlarged, thus biasing the decision toward a forgery. On the other hand, similarities between two points outside the stability regions should be enlarged as well, again biasing the decision toward a forgery. 

\begin{figure*}[!t]
	\centering
	\includegraphics[width=\textwidth]{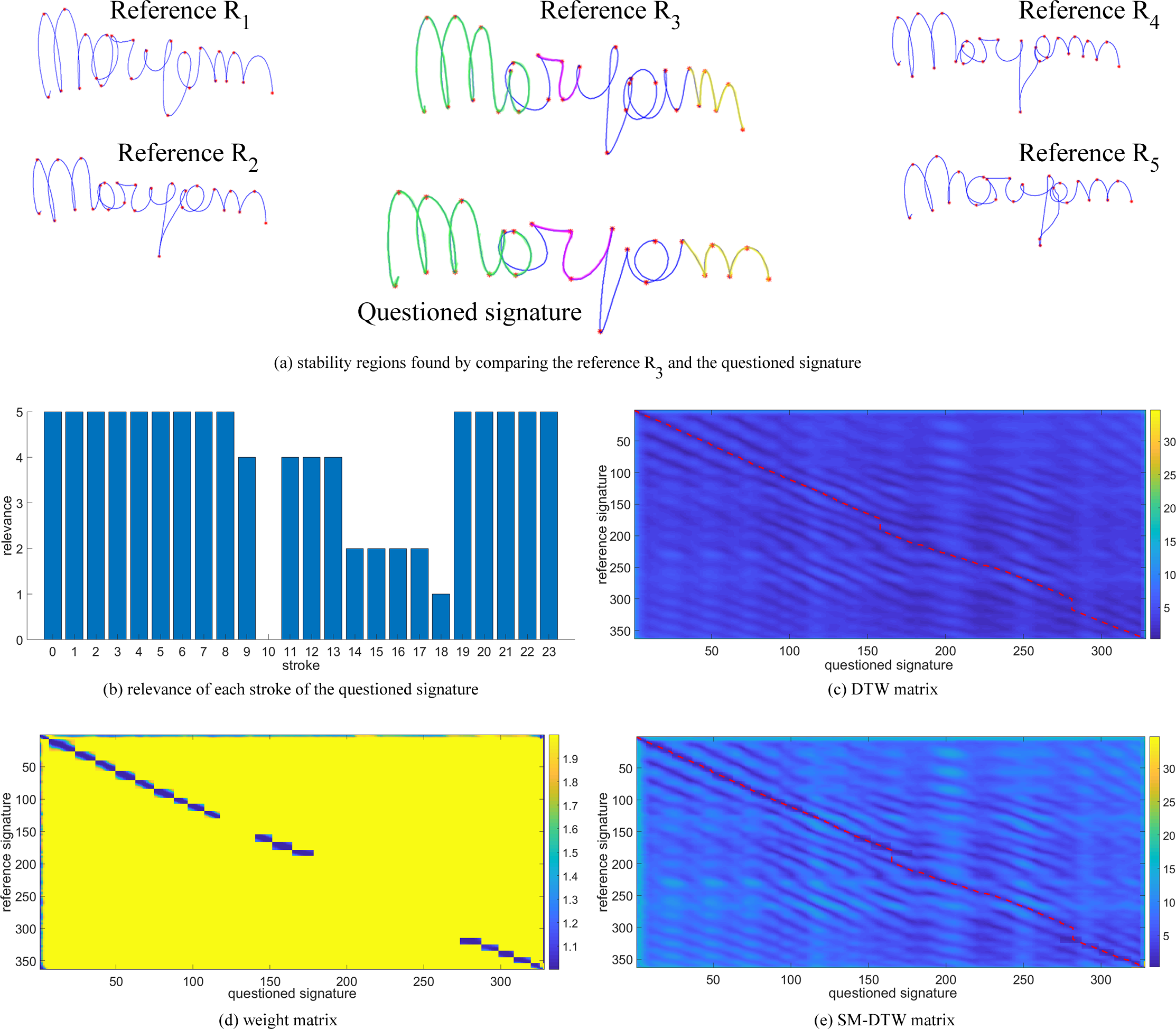}
	\caption{(a) The search for the stability regions. In the top of the figure a set of reference signatures and a questioned signature are shown. The stability regions found between the reference $R_{3}$ and the questioned signature are coloured in green, magenta and yellow. The questioned signature is segmented in 23 strokes. Segmentation points are depicted in red. (b) The relevance of each stroke of the questioned is shown in the histogram. Two regions of the signature are strongly stable (from stroke 0 to 7 and from stroke 19 to stroke 23), whereas stroke 10 doesn't match with any stroke of the reference set. (c) On the right of the histogram, the classical DTW cost matrix is represented. In the bottom part of the image, (d) the weight matrix and (e) the SM-DTW cost matrix are reported. In the DTW and SM-DTW cost matrices is depicted in red the minimum warping path. The visual comparison between the cost matrix of DTW and SM-DTW shows that the modulation due to the stability regions filters out local similarities between points, thus favouring the emergence of the global one.}
	\label{esempio}
\end{figure*}

Figure \ref{esempio} visualises the steps of the process described in this section.

\section{Experiments and results}
\subsection{Datasets}
\label{sec:db}
Two standard and frequently used datasets, MCYT-100 and BiosecureID-SONOF, were adopted for evaluating the performances of the proposed method. 

The MCYT-100 dataset is a corpus of on-line signatures written by 100 users on a tablet. For each user, 25 genuine signatures and 25 skilled forgeries are made available. The skilled forgeries were produced by 5 different users that were allowed to observe the static images of the signature to imitate, and to practice with the copy before producing the valid signatures \cite{MCYT2003}.

The BiosecureID-SONOF dataset is a corpus of on-line, off-line and synthetic off-line signatures produced by 132 subject in four sessions over a six-month period. During each session, 4 genuine signatures and 3 skilled forgeries were acquired for each user. Therefore, for each user 16 genuine signatures and 12 skilled forgeries are made available \cite{2015_PR_onlineOfflineSign_Galbally}. In this work, we use the on-line subcorpus of BiosecureID-SONOF. 

\subsection{Features}\label{featureSec}
As proposed in \cite{fischer2015robust}, each point of the pen-tip trajectory is represented by eight features: the pen position $(x,y)$, the pressure $(p)$, the velocity $(\dot{x},\dot{y})$, the pressure derivative $(\dot{p})$, and the acceleration $(\ddot{x},\ddot{y})$. To compare features vectors by means of an Euclidean distance, the dynamic range of the features was normalized by a z-score as in eq. \ref{eqnorm}.
\begin{equation}\label{eqnorm}
\hat{f_{i}}=\frac{f_{i}-\mu_{f_{i}}}{\sigma_{f_{i}}}
	\end{equation}
where $\mu_{f_{i}}$ and $\sigma_{f_{i}}$ are the mean and the standard deviation computed for the $i^{th}$ feature over all sampling points of the signature.

\subsection{Classification}
To classify a questioned signature as genuine or forged, we evaluated the effects of the two types of normalization applied to the distance computed with eq. \ref{distancepath} (classical DTW, baseline system) and with eq. \ref{distancepathsm} (SM-DTW). Distance normalization is needed in order to compare signatures written by the same subject as well as to identify for all the subjects a common threshold beyond which signatures are classified as forged.

Both the normalizations require preventively to compare each reference signature with all the other ones. The first normalization is defined respect to the ratio between the minimum dissimilarity per subject and the length of its warping path $\mid p\mid_{\text{min}_{\mathbf{R}}}$, as in eq. \ref{minimobl} and \ref{minimoSM}.

\begin{equation}
\label{minimobl}
s^{\text{bl}}_{min}=\frac{\text{min}\, d\left(\overline{{R}_{j}}, \overline{R}_{i}\right)}{\mid p\mid_{\text{min}_{\mathbf{R}}}}
\end{equation}

\begin{equation}
\label{minimoSM}
s^{\text{SM}}_{min}=\frac{\min d^{\text{SM}}\left(\overline{{R}_{j}}, \overline{R}_{i}\right)}{\mid p\mid_{\text{min}_{\mathbf{R}}}}
\end{equation}

The second normalization is defined respect to a value that represents the mean dissimilarity per subject. Such a mean dissimilarity is computed by averaging the distances between a reference and all the others as in eq. \ref{meanS2BL} and \ref{meanS2SM}.

\begin{equation}
\label{meanS2BL}
\mu_{\mathbf{R}}=\frac{1}{\mathit{N}(\mathit{N}-1)}\sum_{i=1}^{\mathit{N}}\sum_{j=1, j\neq i}^{\mathit{N}}\frac{d\left(\overline{R_{i}}, \overline{R}_{j}\right)}{\mid R_{j}\mid}
\end{equation}

\begin{equation}
\label{meanS2SM}
\mu_{\mathbf{R}}^{\text{SM}}=\frac{1}{\mathit{N}(\mathit{N}-1)}\sum_{i=1}^{\mid\mathbf{R}\mid}\sum_{j=1, j\neq i}^{\mathit{N}}\frac{d^{\text{SM}}\left(\overline{R_{i}}, \overline{R}_{j}\right)}{\mid R_{j}\mid}
\end{equation}

When a questioned signature is classified by using the first normalization, the minimum distance between the questioned and the references $\mathbf{R}$ is computed. Such a distance is normalized respect to the length of its warping path $\mid p\mid$. Eventually, the questioned signatures is classified by computing its score as in eq. \ref{scorebaseline} or \ref{scoresm}, depending on whether the distance is computed with the classical DTW or SM-DTW, respectively.

\begin{equation}
\label{scorebaseline}
	s^{\text{bl}}_{1} = \frac{	\min_{{R}_{i}\in \mathbf{R}} d\left(\overline{Q}, \overline{R}_{i}\right)}{\mid p\mid}-s^{\text{bl}}_{min}
\end{equation}

\begin{equation}
	\label{scoresm}
	s^{\text{SM}}_{1} = \frac{\min_{{R}_{i}\in \mathbf{R}} d^{\text{SM}}\left(\overline{Q}, \overline{R}_{i}\right)}{\mid p\mid}-s^{\text{SM}}_{min}
\end{equation}

The second normalization classifies a questioned signature by averaging the distances between this signatures and all the references in $\mathbf{R}$. The distance respect to a reference is normalized respect to $\mid R_{i}\mid$, the length of the reference itself. This choice follows immediately from the concept of stability regions, as it redistribute along the trace the dissimilarity between stability regions, somehow building a model of the signing habit of the subject. The questioned signature is classified by computing its score as in eq. \ref{score2baseline} or \ref{score2sm}, depending on whether the distance is computed with the classical DTW or SM-DTW, respectively.

\begin{equation}
\label{score2baseline}
s^{\text{bl}}_{2} =\text{mean}_{{R}_{i}\in \mathbf{R}}(\frac{d\left(\overline{Q}, \overline{R}_{i}\right)}{\mid R_{i}\mid})-\mu_{\mathbf{R}}
\end{equation}

\begin{equation}
\label{score2sm}
s^{\text{SM}}_{2} = \text{mean}_{{R}_{i}\in \mathbf{R}}(\frac{ d^{\text{SM}}\left(\overline{Q}, \overline{R}_{i}\right)}{\mid R_{i}\mid})-\mu^{\text{SM}}_{\mathbf{R}}
\end{equation}

\subsection{Experimental protocol}
The performance of the baseline and the SM-DTW systems, when both the normalizations were applied, were evaluated in terms of equal error rate (EER), which correspond to the point of the ROC curve where the false acceptance rate equals the false rejection rate. The EER was evaluated in two separate experiments: the Random Forgery (RF) and the Skilled Forgery (SF) experiments. The main difference between these two experiments are the specimens used as impostor signatures. In the Random Forgery (RF) experiment, the impostor signatures are genuine signatures of other signers. It mimics a situation where the impostor has not idea about the signature to fake and try to authenticate in the system with its genuine signature. It is typical in a commercial transaction. Whereas, the impostors in the Skilled Forgery (SF) experiment are people who have previously trained the signature to falsify. This is the most relevant experiment in ASV field.

In both the experiments, the first 5 genuine signatures of each subject were used as reference signatures. In the RF experiment, the first signature of all the other subjects were used as random forgeries. As with regards to this choice, we favour it with respect to randomly selecting them from the available ones and eventually performing cross-validation with different set of references. This is supported by the arguments discussed in \cite{sae2014online}, showing that such a choice better mimics the real scenario, where the reference signatures are acquired during system setup, while the questioned (either genuine or forged) come later, during system operation. Moreover, it has been argued that using as references the signatures that were collected first allows to evaluate the robustness of ASV with respect to the aging of the subjects. Eventually, the argument of repeatability of experiments for better comparison has also been used to favour this choice \cite{das2016multi}. 

As with regards to performance, in the case of MCYT-100, it was evaluated on 9,900 random forgeries, 2,500 skilled forgeries and 2,000 genuine signatures. When the online signatures of BiosecureID-SONOF dataset were used, it was evaluated on 17,292 random forgeries, 1,584 skilled forgeries and 1,452 genuine signatures.

The EER was evaluated by varying the feature set used for representing signatures, as listed in Table \ref{tabFeatures}. It is worth noting that the 15 feature sets were defined by combining exhaustively 4 groups of features: position, pressure, velocity and acceleration.

In \cite{ParzialeFuschetto2013}, it has been studied the connection between the values assumed by the thresholds $th_{LCS}$, $th_{LEN}$ and $th_{GS}$ and the capability of a classifier to discriminate between genuine and forgery signatures when the length of stability regions is used as unique feature. The best performance were obtained when $th_{LCS}$, $th_{LEN}$ and $th_{GS}$ were set equal to 75, 3 and 90 respectively; these are the values used in the experiments presented here. 

\begin{figure*}
	\centering
	\includegraphics[width=\textwidth]{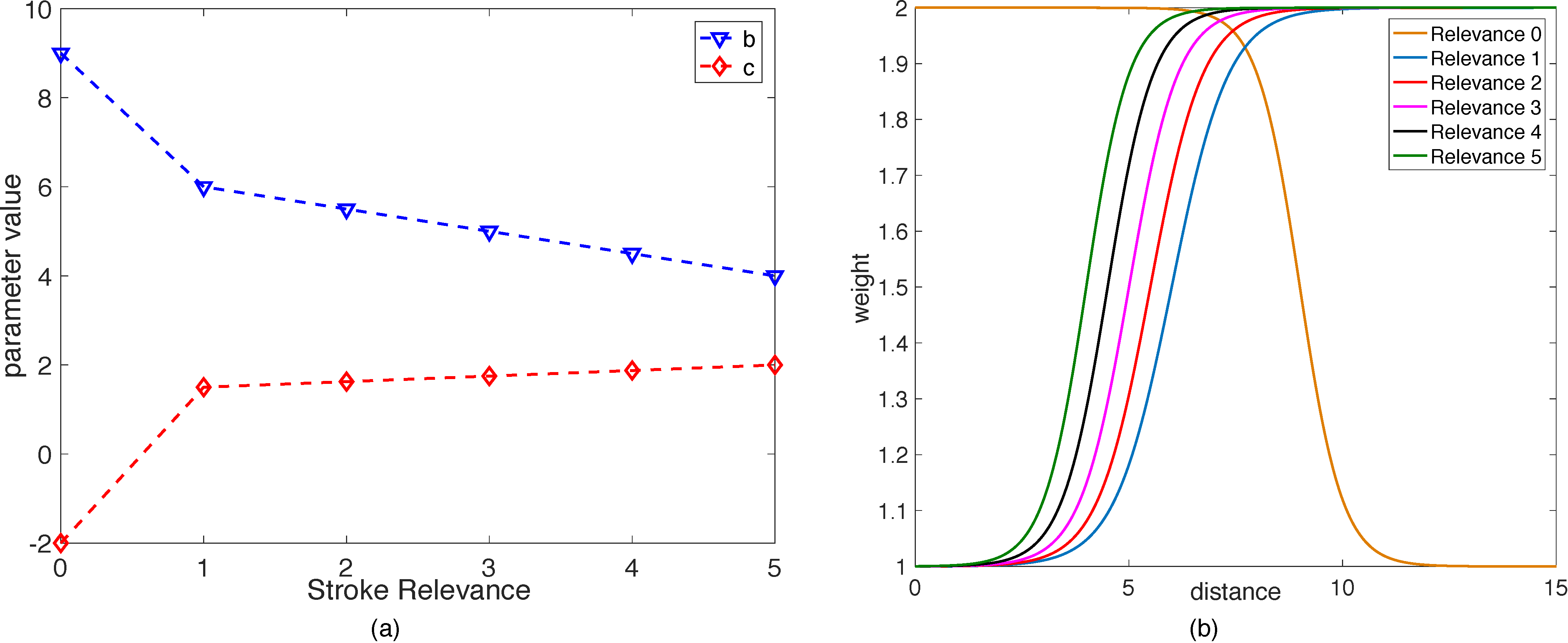}
	\caption{(a) Values of $b$ and $c$ parameters by varying the stroke relevance. (b) Sigmoid functions used for modulating the DTW distance by varying the stroke relevance.}
	\label{figsigmoidi}
\end{figure*}

The values of the parameters $b$ and $c$, used for modulating the cost matrix, were computed by evaluating the performance of SM-DTW algorithm on SVC-task2 dataset \cite{yeung2004svc2004}. We adopted the same experimental protocol described above and we choose the values corresponding to the smallest EER. In particular, the parameters $b_{0}$, $b_{min}$, $b_{max}$, $c_{0}$, $c_{min}$, $c_{max}$ were set equal to 9, 4, 6, -2, 1.5 and 2, respectively. In Figure \ref{figsigmoidi} are depicted the values of the parameters $b$ and $c$ depending on the relevance of the stroke and the resulting weighting functions.

\begin{table}[!t]
	\caption{\label{tabFeatures} Different feature sets evaluated in the experimentation}
	\centering
	\begin{tabular}{|l|c|}
		\hline
		Symbol & Feature Set\\ \hline
		$F_{1}$ & $(x,y,\dot{x},\dot{y},\ddot{x},\ddot{y},p,\dot{p})$\\ \hline
		$F_{2}$ & $(x,y,\ddot{x},\ddot{y},p,\dot{p})$\\ \hline
		$F_{3}$ & $(x,y,\dot{x},\dot{y},p,\dot{p})$\\ \hline
		$F_{4}$ & $(x,y,\dot{x},\dot{y},\ddot{x},\ddot{y})$\\ \hline
		$F_{5}$ & $(\dot{x},\dot{y},\ddot{x},\ddot{y},p,\dot{p})$\\\hline
		$F_{6}$ & $(x,y,p,\dot{p})$ \\ \hline
		$F_{7}$ & $(x,y,\ddot{x},\ddot{y})$ \\ \hline
		$F_{8}$ & $(x,y,\dot{x},\dot{y})$ \\ \hline
		$F_{9}$ & $(\ddot{x},\ddot{y},p,\dot{p})$   \\ \hline
		$F_{10}$ & $(\dot{x},\dot{y},\ddot{x},\ddot{y})$ \\ \hline
		$F_{11}$ & $(\dot{x},\dot{y},p,\dot{p})$\\ \hline
		$F_{12}$ & $(x,y)$\\ \hline
		$F_{13}$ & $(p,\dot{p})$ \\ \hline
		$F_{14}$ & $(\ddot{x},\ddot{y})$ \\ \hline
		$F_{15}$ & $(\dot{x},\dot{y})$\\ \hline			
	\end{tabular}
\end{table}

\subsection{Results}
Table \ref{tableMCYT} and Table \ref{tablebiosecure} report the EER of both the experiments on MCYT-100 and BiosecureID-SONOF datasets, respectively. They report, for each feature set, the perfomance of the baseline and the SM-DTW system: the first six columns of each line reports the results obtained using the first normalization, the last six columns those obtained using the second one.

Let us consider the results on the MCYT-100 dataset when the first type of normalization is used. The EER on skilled forgeries (columns 2, 4 and 6) shows that the SM-DTW system achieves better or comparable performance than the baseline on 12 feature sets, with the largest error reduction (34.20\%) achieved on the feature set $F_{13}$ and the smallest (0.99\%) achieved on $F_{2}$. There are 3 cases when, however, SM-DTW performs worse than baseline, with the largest error increment (10.27\%) reported on $F_{1}$ and the smallest (4.50\%) reported on $F_{7}$. The ERR on random forgeries (columns 1, 3 and 5) shows that SM-DTW performs better than the baseline on 13 feature sets, with the largest error reduction (55.86\%) achieved on $F_{6}$ and the smallest pne (2.51\%) on $F_{4}$, while no reduction was observed on $F_{1}$. There are two cases when SM-DTW performs worse than the baseline, with the largest error increment (12.51\%) reported on $F_{7}$ and the smallest (2.28\%) on $F_{8}$. It is worth noticing that the best performance on both kind of forgeries is achieved on the same feature set, namely $F_{1}$ and $F_{5}$ for the baseline and the SM-DTW, respectively, with the baseline system offering slightly better performance on the skilled forgeries. 

When the second type of normalization is considered, the ERR on skilled forgeries (columns 8, 10 and 12) shows that SM-DTW systems performs better than the baseline on 14 feature sets, with the largest error reduction (26.87\%) achieved on $F_{13}$ and the smallest one (4.36\%) on $F_{3}$, while no reduction was achieved on $F_{8}$ and $F_{11}$. On the features set $F_{7}$, the baseline system performs better than the SM-DTW, for which a small increment (1.27\%) of the EER was reported.

The ERR on random forgeries (columns 7, 9 and 11) shows that SM-DTW outperforms the baseline on any feature set, with the largest error reduction (34.81\%) achieved on $F_{14}$ and the smallest (8.33\%) on $F_{8}$. It is worth noticing that in this case there is not a single features set on which the systems achieve the best performance on both random and skilled forgeries. Considering that the SF experiment is the the most relevant in ASV field, the best performance of the baseline system (3.45\%) is achieved on the feature set $F_{2}$, with an ERR of 1.30\% on the random forgeries. The SM-DTW achieves its best performance (3.09\%) on $F_{5}$, with an error reduction of 10.43\% with respect to the best performance of the baseline, and the same EER on random forgeries. It is worth pointing out that the SM-DTW system achieves its best performance on the same feature set, whichever normalization is used. On the contrary, the baseline achieves its best performance on $F_{1}$ or $F_{2}$, depending on which normalization is used.

In the case of the BiosecureID-SONOF dataset, when the first normalization is used the EER on skilled forgeries (columns 2, 4 and 6) shows that the SM-DTW system achieves better performance than the baseline on 14 feature sets, with the largest error reduction (42.95\%) achieved on the feature set $F_{13}$ and the smallest (1.31\%) achieved on $F_{15}$. On $F_{1}$, however, SM-DTW performs worse than baseline, reporting an error increment of 1.71\%. The ERR on random forgeries (columns 1, 3 and 5) shows that SM-DTW outperforms the baseline on every feature set, with the largest error reduction (65.38\%) achieved on $F_{5}$ and the smallest (14.37\%) on $F_{4}$. It is worth noticing that there is not a single feature set on which the systems achieve the best performance on both random and skilled forgeries. Again, considering the SF experiment as the most relevant one, the best performance of the baseline (3.85\%) is achieved on the feature set $F_{1}$, with an ERR of 0.69\% on the random forgeries, while the SM-DTW achieves its best performance (3.39\%) on $F_{5}$, with an ERR of 0.33\% on the random forgeries. Comparing the best performance, the SM-DTW achieves an error reduction with respect to the baseline of 11.95\% and 52.17\% for skilled and random forgeries, respectively.

When the second type of normalization is used, the ERR on skilled forgeries (columns 8, 10 and 12) shows that SM-DTW systems performs better than the baseline on 13 feature sets, with the largest error reduction (20.61\%) achieved on $F_{2}$ and the smallest one (2.54\%) on $F_{8}$, while an increment of the EER was reported on two feature sets, the largest (17.94\%) on $F_{10}$ and the smallest (13.14\%) on $F_{15}$. 

The ERR on random forgeries (columns 7, 9 and 11) shows that SM-DTW outperforms the baseline on any feature set, with the largest error reduction (42.89\%) achieved on $F_{8}$ and the smallest (18.16\%) on $F_{15}$. It is worth noticing that also in this case there is not a single feature set on which the systems achieve the best performance on both random and skilled forgeries. Considering the SF experiment as the most relevant one, the best performance of the baseline (1.65\%) is achieved on $F_{10}$, with an ERR of 1.44\% on the random forgeries, while the SM-DTW achieves its best performance (1.45\%) on $F_{5}$, with an error reduction of 12.12\%, and an EER of 1.17\%, with an error reduction of 18.75\%. It is worth pointing out that, as for the MYCT-100 dataset, the SM-DTW system achieves its best performance on $F_{5}$, whichever normalization is used. On the contrary, the baseline system achieves its best performance on $F_{1}$ or $F_{10}$, depending on which normalization is used.

\begin{table*}[!t]
	\caption{\label{tableMCYT}EER values on MCYT-100 dataset}
	\centering
	\begin{tabular}{|c||c|c||c|c||c|c||c|c||c|c||c|c|}
		\hline
		& \multicolumn{6}{ c||}{ $s_{1}$} & \multicolumn{6}{c|}{$s_{2}$}\\\hline
		& \multicolumn{2}{c||}{DTW}& \multicolumn{2}{c||}{SM-DTW}& \multicolumn{2}{ c||}{Improvement}& \multicolumn{2}{c||}{Baseline} & \multicolumn{2}{c||}{SM-DTW}& \multicolumn{2}{c|}{Improvement}\\ \hline
		Features& RF & SF & RF & SF& $\Delta\text{RF}$&$\Delta\text{SF}$& RF & SF & RF & SF& $\Delta\text{RF}$&$ \Delta\text{SF}$\\\hline
		$F_{1}$& 0.30& 4.09& 0.30& 4.51& 0.00\%& -10.27\%& 1.35& 3.49& 1.15& 3.25& 14.87\%& 7.02\%\\ \hline
		$F_{2}$& 0.40& 4.56& 0.30& 4.51& 25.00\%& 0.99\%& 1.30& 3.45& 1.15& 3.25& 11.58\%& 5.81\%\\ \hline
		$F_{3}$& 0.40& 5.25& 0.30& 4.96& 25.00\%& 5.53\%& 1.20& 3.56& 1.05& 3.40& 12.55\%& 4.36\%\\ \hline
		$F_{4}$& 0.40& 4.80& 0.39& 4.60& 2.51\%& 4.17\%& 1.40& 3.96& 1.10& 3.76& 21.50\%& 5.06\%\\ \hline
		$F_{5}$& 0.35& 4.20& 0.30& 4.45& 14.30\%& -5.83\%& 1.70& 3.45& 1.30& 3.09& 23.37\%& 10.30\%\\ \hline
		$F_{6}$& 0.90& 7.65& 0.40& 5.91& 55.86\%& 22.69\%& 1.55& 4.05& 1.20& 3.71& 22.40\%& 8.28\%\\ \hline
		$F_{7}$& 0.40& 4.45& 0.45& 4.65& -12.51\%& -4.50\%& 1.30& 3.56& 1.01& 3.60& 22.39\%& -1.27\%\\ \hline
		$F_{8}$& 0.44& 6.00& 0.45& 5.25& -2.28\%& 12.58\%& 1.15& 3.96& 1.05& 3.96& 8.33\%& 0.00\%\\ \hline
		$F_{9}$& 0.85& 5.56& 0.50& 4.56& 41.42\%& 18.00\%& 2.14& 3.51& 1.55& 3.31& 27.70\%& 5.70\%\\ \hline
		$F_{10}$& 0.59& 5.20& 0.41& 4.60& 30.49\%& 11.54\%& 1.95& 4.05& 1.36& 3.76& 30.15\%& 7.17\%\\ \hline
		$F_{11}$& 0.65& 5.96& 0.36& 5.00& 44.96\%& 16.04\%& 1.70& 3.56& 1.25& 3.56& 26.33\%& 0.00\%\\ \hline
		$F_{12}$& 0.90& 9.65& 0.75& 7.76& 16.76\%& 19.60\%& 1.25& 5.16& 1.10& 4.76& 12.05\%& 7.76\%\\ \hline
		$F_{13}$& 7.60& 20.60& 4.31& 13.56& 43.25\%& 34.20\%& 6.25& 8.36& 4.20& 6.11& 32.80\%& 26.87\%\\ \hline
		$F_{14}$& 0.80& 5.65& 0.70& 4.96& 12.58\%& 12.22\%& 2.15& 4.16& 1.40& 3.80& 34.81\%& 8.54\%\\ \hline
		$F_{15}$& 0.65& 5.96& 0.40& 5.51& 37.98\%& 7.47\%& 1.26& 4.25& 1.05& 3.91& 16.74\%& 7.89\%\\ \hline
		
	\end{tabular}
\end{table*}

\begin{table*}[!t]
	\caption{\label{tablebiosecure}EER values on BiosecureID-SONOF dataset}
	\centering
	\begin{tabular}{|c||c|c||c|c||c|c||c|c||c|c||c|c|}
		\hline
		& \multicolumn{6}{ c||}{ $s_{1}$} & \multicolumn{6}{c|}{$s_{2}$}\\\hline
		& \multicolumn{2}{c||}{DTW}& \multicolumn{2}{c||}{SM-DTW}& \multicolumn{2}{ c||}{Improvement}& \multicolumn{2}{c||}{Baseline} & \multicolumn{2}{c||}{SM-DTW}& \multicolumn{2}{c|}{Improvement}\\ \hline
		Features& RF & SF & RF & SF& $\Delta\text{RF}$&$\Delta\text{SF}$& RF & SF & RF & SF& $\Delta\text{RF}$&$ \Delta\text{SF}$\\\hline
		$F_{1}$& 0,69& 3,85& 0,33& 3,92& 51,89\%& -1,71\%& 1,52& 2,01& 1,17& 1,65& 22,82\%& 18,00\%\\\hline
		$F_{2}$& 0,62& 4,91& 0,33& 3,92& 46,52\%& 20,12\%& 1,86& 2,07& 1,17& 1,65& 37,15\%& 20,61\%\\\hline
		$F_{3}$& 0,69& 5,11& 0,28& 4,61& 59,83\%& 9,67\%& 1,52& 2,21& 1,17& 1,85& 23,25\%& 16,38\%\\\hline
		$F_{4}$& 0,48& 3,99& 0,41& 3,66& 14,37\%& 8,28\%& 1,30& 2,14& 1,03& 1,94& 20,75\%& 9,25\%\\\hline
		$F_{5}$& 0,97& 3,85& 0,33& 3,39& 65,38\%& 11,99\%& 1,79& 1,71& 1,17& 1,45& 34,62\%& 15,41\%\\\hline
		$F_{6}$& 1,16& 8,00& 0,63& 5,86& 45,88\%& 26,75\%& 2,43& 2,60& 1,58& 2,14& 34,80\%& 17,75\%\\\hline
		$F_{7}$& 0,69& 5,11& 0,47& 4,28& 31,52\%& 16,13\%& 1,79& 2,21& 1,24& 2,14& 30,87\%& 2,99\%\\\hline
		$F_{8}$& 0,55& 5,99& 0,47& 5,17& 14,67\%& 13,74\%& 1,52& 2,60& 1,16& 2,54& 23,77\%& 2,54\%\\\hline
		$F_{9}$& 1,51& 5,86& 0,69& 4,05& 54,10\%& 30,89\%& 2,41& 1,71& 1,38& 1,52& 42,89\%& 11,56\%\\\hline
		$F_{10}$& 0,89& 4,05& 0,55& 3,92& 38,51\%& 3,26\%& 1,44& 1,65& 0,97& 1,94& 33,13\%& -17,94\%\\\hline
		$F_{11}$& 1,10& 5,80& 0,48& 4,61& 56,40\%& 20,44\%& 2,00& 1,94& 1,24& 1,58& 37,81\%& 18,61\%\\\hline
		$F_{12}$& 1,73& 11,99& 0,97& 9,09& 44,18\%& 24,17\%& 2,34& 3,92& 1,80& 3,79& 23,25\%& 3,37\%\\\hline
		$F_{13}$& 7,56& 18,94& 3,86& 10,80& 49,01\%& 42,95\%& 6,19& 4,05& 4,41& 3,52& 28,77\%& 13,03\%\\\hline
		$F_{14}$& 1,38& 4,81& 0,76& 4,05& 44,98\%& 15,75\%& 1,31& 1,71& 0,97& 1,65& 26,21\%& 3,85\%\\\hline
		$F_{15}$& 0,77& 5,04& 0,61& 4,97& 20,99\%& 1,31\%& 1,09& 2,01& 0,90& 2,27& 18,16\%& -13,14\%\\\hline
	\end{tabular}
\end{table*}

\subsection{Comparison with the State of the art}
To put in context the performing properties of the SM-DTW algorithm in signature verification, we selected the most performing and recently published methods using the same datasets for performance evaluation. In Table~\ref{tab1}, we compare the performance of seven state of the art methods on the MCYT-100 dataset, sorted according to the performance on the SF experiment. The first one adopts a representation of the signatures in terms of vector quantization \cite{sharma2016enhanced}, while the second method carries out a fusion that use the information from the cost matrix and warping path alignments \cite{sharma2017exploration}. The third and fifth methods combine the DTW with Gaussian Mixture Model features, with the third one adopting a fusion strategy to combine information provided by both sources \cite{sharma2017novel}, and the fifth exploiting only GMM-related features for the final classification. The fourth method combines DTW and $\Sigma \Lambda$-based features \cite{fischer2017signature}. The sixth method uses a histogram distribution of the features and a Manhattan distance as classifier \cite{sae2014online}, whereas the last one adopts a symbolic representation of the signatures \cite{guru2017interval}.

When both the RF and SF experiments were performed, the results show that SM-DTW performs better than the other methods in the SF experiment, but worse than two of them in the RF experiment. As a whole, they show that DTW-based methods outperform the other ones in both the experiments. More specifically, we note that in the SF experiment SM-DTW performs better of methods that, in addition to the information provided by the DTW, use other features to characterise the signing habits of the subjects. In our opinion, the comparison with those methods is of particular interest, because it shows that introducing the stability regions for characterising the signature of a subject leads to comparable or better performance than those obtained by using much more sophisticated feature set and classifiers. When only the SF experiment was performed, SM-DTW is outperformed by methods that combine the results obtained when using DTW-based information with other sources of information, as it is well represented by the two methods exploiting GMM-based features \cite{sharma2017novel}, which shows a boost in performance when fusion is adopted. 

In Table~\ref{tab2}, we show the performance on the BiosecureID-SONOF dataset. The top system is the one presented in this paper. The second system \cite{Ferrer2017} is based on the Manhattan distance between histograms of different measures of the signature under comparison. The third classifier \cite{Ferrer2017} is based on DTW with Euclidean distance and parameter vector. Next, the fourth method fuses on-line signatures and synthetic off-line signatures, being the DTW used in the dynamic verification and a Support Vector Matching classifier during the static signature verification. 
 
It is important noticing that the quantitative data reported in Table~\ref{tab1} and Table~\ref{tab2} have not been obtained in the same conditions, as we have already discussed when presenting our experimental protocol, so that a direct comparison in terms of EER may be misleading. Nonetheless, they do show that SM-DTW adopting as features velocity, acceleration and pressure ($F_{5}$) and whose score has been normalised by using the second methods ($s_{2}^{\text{SM}}$) is outperformed only by methods that exploit fusion strategy of multiple source of information for making the final decision.

\begin{table}[!t]
	\caption{\label{tab1} Related works on MCYT-100. Results in terms of EER\,\%}
	\centering
	\scriptsize
	\begin{tabular}{|l|c|c|}
		\hline
		Method & RF & SF \\ \hline \hline
		VQ+DTW (fusion, user threshold) \cite{sharma2016enhanced} & - & 1.55 \\ \hline
		WP+BL DTW fusion \cite{sharma2017exploration} & - & 2.76 \\ \hline
		GMM+DTW with Fusion \cite{sharma2017novel} & - & 3.05 \\ \Xhline{4\arrayrulewidth}
		\textbf{Proposed Method - SM-DTW $s_{2}^{SM}$ $F_{5}$} &1.30&3.09 \\\hline
		$\Sigma \Lambda$ + DTW \cite{fischer2017signature} & 1.01 & 3.56 \\ \hline
		GMM+DTW \cite{sharma2017novel} & - & 4.00 \\ \hline
		Histogram + Manhattan \cite{sae2014online} & 1.15 & 4.02 \\ \hline
		Symbolic representation \cite{guru2017interval} & 2.40  & 5.70 \\ \hline
	\end{tabular}
\end{table}

\begin{table}[!t]
	\caption{\label{tab2} Related works on BiosecureID-SONOF. Results in terms of EER\,\%}
	\centering
	\scriptsize
	\begin{tabular}{|l|c|c|}
		\hline
		Method & RF & SF \\ \hline \hline
		\textbf{Proposed Method - SM-DTW $s_{2}^{SM}$ $F_{5}$} &1.17&1.45\\\hline
		Histogram based and Manhattan distance \cite{Ferrer2017} & 1.16 & 1.98 \\ \hline 
		Function based and DTW distance \cite{Ferrer2017} & 0.23 & 3.08  \\ \hline
		ASV fusion (On-line + Off-line) \cite{2015_PR_onlineOfflineSign_Galbally} & 0.63 & 5.09  \\ \hline  
   
	\end{tabular}
\end{table}

\section{Conclusions}
The main purpose of this work has been that of providing experimental evidence to support our conjecture that, as the signing habits of a subject have been encoded in the motor plan that has been learned, during multiple executions of the signature, the movements corresponding to the command stored into the motor plan should produce very similar trajectories. It may be possible, however, that the complex movements corresponding to a signature have not been completely learned, and therefore the motor plan encodes only parts of the movements, while others are computed on the fly. As a consequence, the trajectory of a signature can be thought as composed of stability regions, originated by the execution of the motor plan, and other regions, originated by the execution computed on the fly. Such stability regions represent the motor plan of the subject, and convey the most relevant information about the signing habits, and should therefore given more importance during the verification.

The data reported in Table \ref{tableMCYT} and Table \ref{tablebiosecure} show that the baseline system best performance is achieved by using \textit{different} feature sets, depending on which dataset is used. On the contrary, SM-DTW achieves better performance by using the \textit{same} feature set, independently of which dataset is used. This means that the stability regions we have introduced seems to capture distinctive aspects of the \textit{process} of signing, i.e. its motor control plan, rather those of a population of \textit{signatures}. The data also show that the SM-DTW achieves its best results when considering the features describing the dynamic of the movement, but not the one describing its trajectory ($F_{5}$). Furthermore, they show that exploiting shape information by the stability regions provides better results than when it is used as any other one ($F_{1}$). This finding confirms that shape information is better exploited by the stability regions rather than by combining trajectory and dynamic aspects of signatures. 
All together the results reported in the previous section support our claim that the stability regions encode the signature motor plan and therefore should be given more importance during signature verification.

Eventually, the results reported in Table ~\ref{tab1} and ~\ref{tab2} show that the SM-DTW system adopting the normalization ($s_{2}^{\text{SM}}$) exhibits EER comparable or better than those of top performing systems based on DTW, and that the SM-DTW is outperformed only when a fusion strategy is used to combine multiple sources of information. 

So far, the results reported in the previous section have been obtained by using the weights defined in subsection 3.4. The next step would be that of computing their values by some optimization technique, to find the upper bound for the performance of the proposed algorithm.

Eventually, our definition of stability regions is based only on the shape of the trajectory. In our future investigations, we will consider exploiting at a larger extent the kinematics of the movement for extracting the strokes embedded into the motor plan. As, according to the findings about handwriting learning and execution mentioned in the Introduction, it is very unlikely that a forger can reproduce both the shape and the dynamics of the signer, we believe that using such information may lead to extracting more relevant stability regions, or even to establish which ones are the features that allow for the best recovery of the motor plan, and therefore to further improvements of the performance with respect to the baseline system.

\section*{Acknowledgments}
This study was partially funded by the grant “Bando PRIN2015 - Progetto HAND” from the Italian Ministero dell'Istruzione, dell'Universit\`{a} e della Ricerca, the first IAPR Research Scholarship Program, the Spanish government's MIMECO TEC2016-77791-C4-1-R research project and European Union FEDER program/funds. 


\bibliographystyle{model2-names.bst}
\bibliography{refs}

\end{document}